\documentclass{bmvc2k}

\title{HairGS: Hair Strand Reconstruction based on 3D Gaussian Splatting}

\addauthor{Yimin Pan}{yimin.pan@tum.de}{}
\addauthor{Matthias Niessner}{niessner@tum.de}{}
\addauthor{Tobias Kirschstein}{tobias.kirschstein@tum.de}{}

\addinstitution{
 The Visual Computing \& Artificial Intelligence Group\\
 Technical University of Munich\\
 Munich, DE
}

\runninghead{Pan et.al.}{Hair Strand Reconstruction based on 3D Gaussian Splatting}

\def\etal{\emph{et al}\bmvaOneDot}

\begin{document}
\maketitle

\begin{figure}[h]
	\label{fig:teaser}
	\centering
	\includegraphics[width=1.0\textwidth]{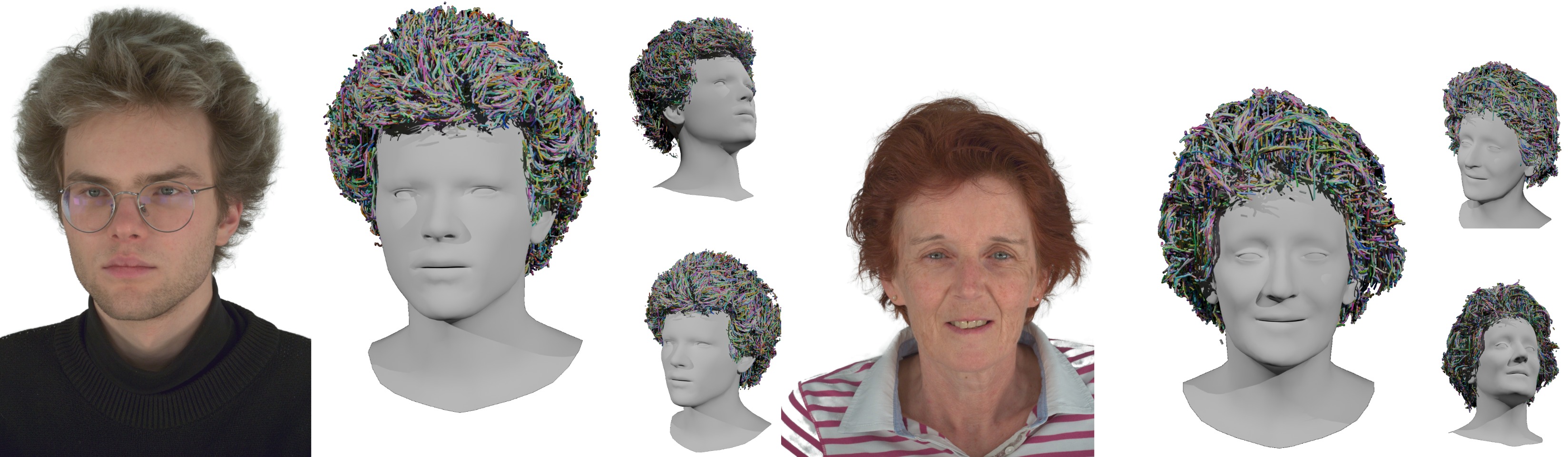}
	\caption{\textbf{Our method reconstructs realistic hair strands across a wide range of hairstyles.} Hair strands are rendered with different colors to facilitate visualization.}
\end{figure}

\begin{abstract}
	Human hair reconstruction is a challenging problem in computer vision, with growing importance for applications in virtual reality and digital human modeling. Recent advances in 3D Gaussians Splatting (3DGS) provide efficient and explicit scene representations that naturally align with the structure of hair strands. In this work, we extend the 3DGS framework to enable strand-level hair geometry reconstruction from multi-view images. Our multi-stage pipeline first reconstructs detailed hair geometry using a differentiable Gaussian rasterizer, then merges individual Gaussian segments into coherent strands through a novel merging scheme, and finally refines and grows the strands under photometric supervision.

	While existing methods typically evaluate reconstruction quality at the geometric level, they often neglect the connectivity and topology of hair strands. To address this, we propose a new evaluation metric that serves as a proxy for assessing topological accuracy in strand reconstruction. Extensive experiments on both synthetic and real-world datasets demonstrate that our method robustly handles a wide range of hairstyles and achieves efficient reconstruction, typically completing within one hour.

	The project page can be found at: \url{https://yimin-pan.github.io/hair-gs/}
\end{abstract}

\section{Introduction}
\label{sec:intro}

Realistic human avatar modeling plays a crucial role in the gaming and virtual reality (VR) industry. Since the introduction of 3D Morphable Models (3DMM) by Blanz and Vetter~\cite{3dmm}, there has been a growing trend toward controllable head modeling. Recent advances in neural representations have further enhanced this field, enabling high-fidelity facial reconstructions~\cite{lombardi2021mixturevolumetricprimitivesefficient,kirschstein2023nersemble,nofa,qian2023gaussianavatars}.

Despite these advancements, most head reconstruction methods focus primarily on facial geometry, often negleting the crucial role hair plays in defining personal identity. Traditional hair modeling tools are highly skill-intensive and time-consuming, as achieving realistic results requires the manual creation of a vast number of individual strands. This challenge has driven a shift toward image-based optimization techniques that aim to automate the hair reconstruction process.

However, hair reconstruction remains a challenging task due to the complex geometry of strands, frequent occlusions, and the limitations of conventional keypoint and multi-view stereo (MVS) techniques~\cite{Colmap, sift}. Data-driven approaches~\cite{Neural-Strands, Neural-Haircut, zakharov2024gh} leverage learned priors to infer hair structure from images, enabling the recovery of even occluded inner volumes. Yet, their generalization capabilities are often limited by the scarcity of high-quality 3D data.

Recently, Gaussian Splatting~\cite{GS} has emerged as a popular method for efficient and accurate scene reconstruction. However, its general-purpose design is not directly applicable for hair modeling. When applied directly, it results in numerous uncontrollable and disconnected Gaussians, and struggles particularly with modeling curly strands.

To address these limitations, we propose a novel multi-stage pipeline that adapts the 3DGS framework with constraints specifically tailored for hair strand reconstruction. In the first stage, we employ a differentiable rasterizer together with adaptive densification to obtain a dense and detailed representation of visible hair geometry. This step is essential, as successful merging relies on comprehensive geometry recovery to identify most suitable candidates. We then introduce a merging scheme based on distance and angle heuristics, enabling the combination of separated hair segments into longer strands. In the final stage, we refine the joint positions using a combination of photometric losses and smoothness regularization. Extensive quantitative and qualitative evaluations demonstrate that our method is general enough to reconstruct a wide variety of hairstyle and substantially faster than most data-driven approaches.
Our main contributions can be summarized as follows:
\begin{itemize}
	\item An efficient multi-stage optimization pipeline and a set of losses to supervise hair directions and prevent the formation of sharp angles.
	\item A merging scheme to combine short strands into longer ones using geometric heuristics.
	\item A novel metric for quantitative evaluation of topological strand accuracy.
\end{itemize}

\section{Related Work}
\label{sec:related_work}

\noindent\textbf{Optimization-Based Hair Reconstruction.} Early approaches to hair strand reconstruction typically rely on Structure from Motion (SfM) to estimate hair geometry. For instance, Paris \etal~\cite{Gabor} formulated the optimization problem using 2D orientation fields derived from Gabor filters as constraints. Luo \etal~\cite{luo12} introduced a Markov Random Field (MRF) optimization to generate oriented point clouds, which were further refined with Poisson surface reconstruction and extended through visual hull carving and orientation-based refinement~\cite{luo13}.
Despite these advances, such methods often produce sparse and noisy results due to the planar neighborhood assumptions of stereo vision~\cite{sift}, which do not hold for hair strands. Works like \cite{BARTOLI05, Bay05, Usumezbas16} addressed the reconstructing problem using properties from line primitives. Nam \etal~\cite{Strand-Accurate} similarly tackled the challenge using LP-MVS, incorporating line constraints within a multi-phase optimization to recover accurate strands. Nevertheless, estimating inner hair volumes from images remains impractical due to heavy occlusion. More recently, ~\cite{Dr-Hair} address this issue by introducing guide strands to interpolate and reconstruct full hair volumes.

\noindent\textbf{Learning-Based Hair Reconstruction.} Learning-based approaches have progressed rapidly with the release of public datasets such as~\cite{usc-hairsalon, Yuksel2009}, enabling model pretraining on synthetic hair data. Recent methods employ neural architectures to synthesize 3D hair strands from images. For example,~\cite{HairNet} uses a CNN-based autoencoder to infer strands from a single image, while~\cite{Hair-GAN} generates volumetric hair fields using GANs. More recent works~\cite{NeuralHDHair, Neural-Strands, Neural-Haircut} leverage implicit neural representations, texture-based encodings, and diffusion priors to further improve fidelity and realism. Nevertheless, these methods are usually limited by the scarcity of high-quality training data, often resulting in reduced expressiveness and oversmoothing, as shown in the experiments (see Figure~\ref{fig:nersemble}).

\noindent\textbf{Radiance Fields and Volumetric Rendering.} The introduction of Neural Radiance Fields (NeRF)~\cite{NeRF} and volumetric rendering has greatly advanced photorealistic human avatar modeling, enabling flexible representation of line-like structures such as hair, which are difficult to capture with traditional mesh-based methods. More recently, 3DGS~\cite{GS} has emerged as an explicit and efficient alternative, representing scenes with 3D Gaussians that naturally align with the structure of hair strands. While prior works~\cite{GroomCap, GS-Hair, zakharov2024gh} typically use 3DGS for refining coarse geometry, our approach leverages 3DGS from the begining to recover precise hair segments, which are then merged into complete strands.

\section{Method}
\label{sec:method}

\begin{figure}
	\centering
	\includegraphics[width=1.0\textwidth]{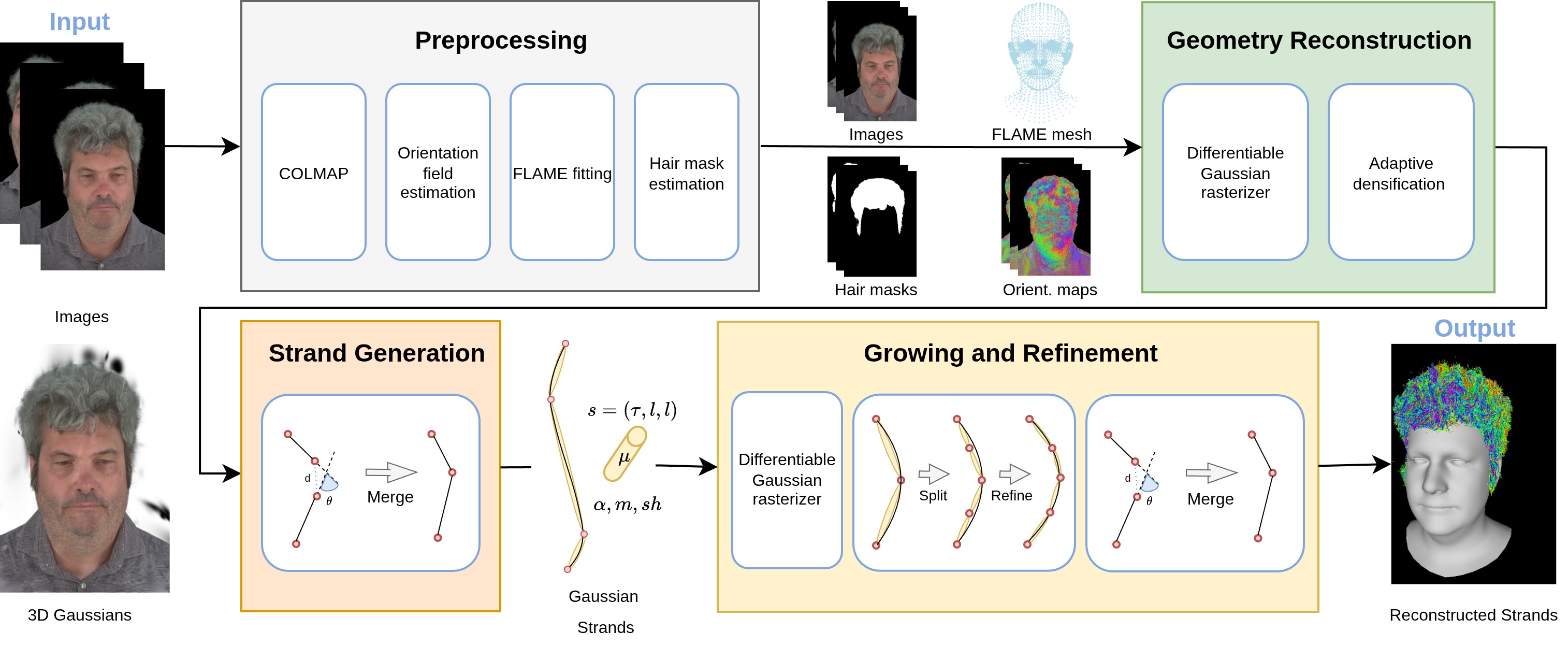}
	\caption{\textbf{An overview of the full pipeline.} The process begins with image preprocessing and geometry reconstruction using 3D Gaussians. Hair strands are created by merging Gaussians, followed by a refinement stage to optimize joint positions and the topology.}
	\label{fig:pipeline}
\end{figure}

Given a set of multi-view images of a static scene, our method reconstructs an accurate 3D hair model of the subject in the form of polylines. Similar to~\cite{GS}, we begin by preprocessing the images using COLMAP~\cite{Colmap} to obtain camera poses and a sparse point cloud. However, due to the sensitivity of 3DGS to initialization, we use the vertices obtained from fitting a FLAME model~\cite{FLAME}. Additionally, we estimate 2D orientation maps and corresponding confidence maps using Gabor filters~\cite{Gabor}, and extract hair segmentation masks using off-the-shelf methods~\cite{MODNet, CDGNet}. The complete pipeline is illustrated in Figure~\ref{fig:pipeline}, which consists of three main stages: geometry reconstruction (Section~\ref{sec:stage_1}), strand generation (Section~\ref{sec:stage_2}), and final refinement (Section~\ref{sec:stage_3}). Unlike other concurrent works, our approach leverages the flexibility of unconstrained Gaussian splatting to recover as many hair segments as possible, which are then merged into coherent strands.

The final output is a set of strands $S = \{S_1, \dots, S_n\}$, each comprising a variable number of linked 3D points $S_i = \{ p_{1}, \dots, p_{m} \}, p_{j} \in \mathbf{R}^3$. In contrast to most existing methods~\cite{Neural-Strands, Neural-Haircut, Strand-Accurate, Strand-Integration, GroomCap}, we do not explicitly specify the number of points per strand, as this is typically unknown and highly dependent on the shape and complexity of the curve being modeled. Instead, we rely on the optimization process and gradient feedback to dynamically determine whether the current number of points is sufficient or if further densification is required.

\subsection{Stage I: Geometry Reconstruction}
\label{sec:stage_1}
\begin{figure}
	\centering
	\includegraphics[width=1.0\textwidth]{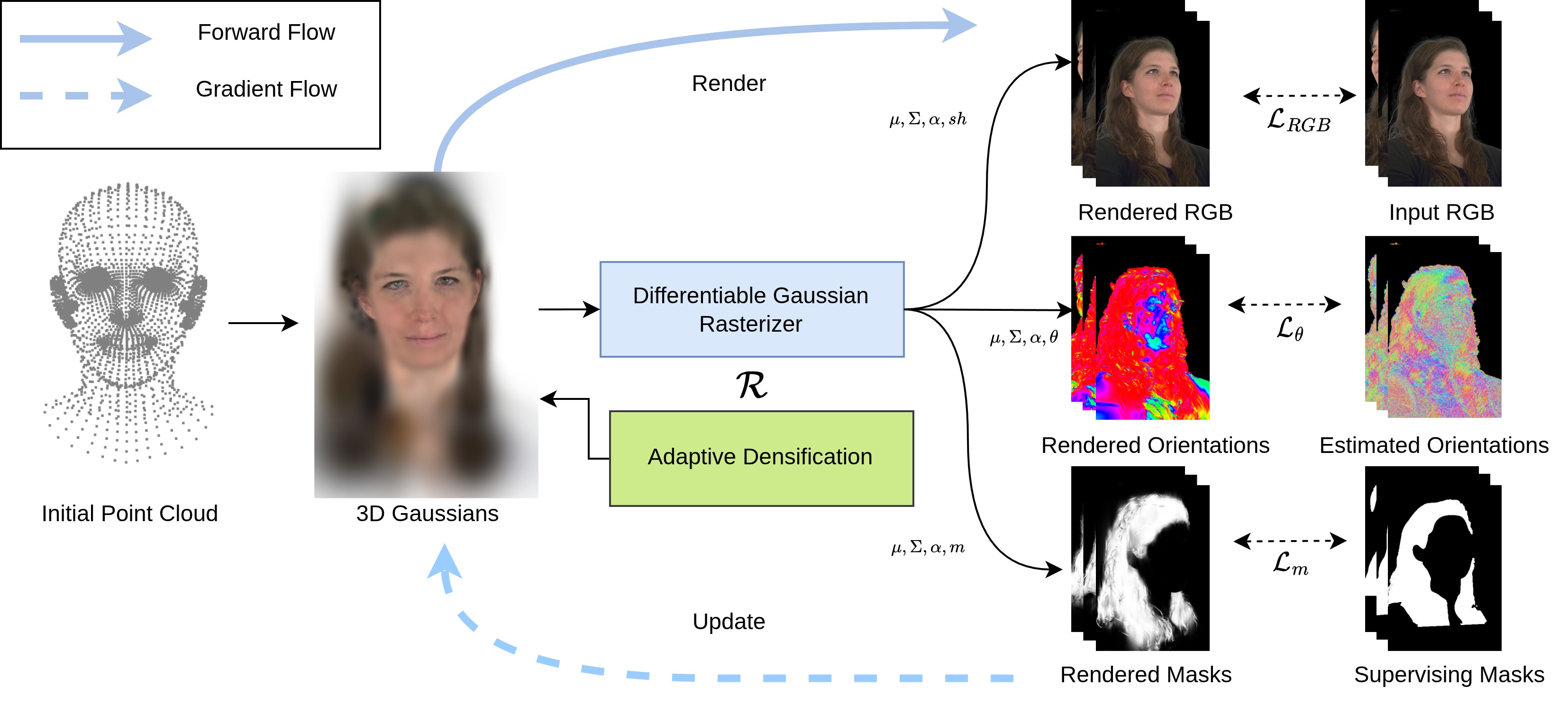}
	\caption{\textbf{Overview of the first-stage process}. From an initial point cloud, 3D Gaussians are optimized using the differentiable rasterizer with the supervision of a combination of RGB, orientation, and mask losses ($\mathcal{L}_{RGB}$, $\mathcal{L}_\theta$, $\mathcal{L}_m$). The densification process populates Gaussians in areas with missing geometry.}
	\label{fig:first_stage}
\end{figure}

The first stage of our pipeline reconstructs hair strand geometry by optimizing photometric losses within an analysis-by-synthesis framework, as shown in Figure~\ref{fig:first_stage}. Following the original 3DGS, we initialize anisotropic 3D Gaussians from the input point cloud, each parameterized by a position mean $\mu\in\mathbf{R}^3$, per-axis scale $s\in\mathbf{R}^3$, a rotation matrix $R$ (as a quaternion) and an opacity value $0 \leq \sigma(\alpha) \leq 1$, with $\sigma(x)$ denoting the sigmoid function. We introduce an additional mask value $0 \leq \sigma(m) \leq 1$, which is crucial for restricting the merging process to valid hair regions and for filtering out non-hair elements in the final output. The corresponding Gaussian distribution in world space is:
\begin{equation}
	G(x) = e^{-\frac{1}{2}(x)^T\Sigma^{-1}(x)}, \quad \Sigma=RS S^{T}R^{T}
	\label{eq:gaussian}
\end{equation}
To focus on geometry rather than appearance, we represent the radiance field using spherical harmonics (SH) coefficients of degree 0, thereby avoiding view-dependent color effects.

We optimize the Gaussian parameters using the differentiable rasterizer $\mathcal{R}$ from~\cite{GS}. The optimization is driven by a combination of photometric losses, $\mathcal{L}_1$ and $\mathcal{L}_{DSSIM}$, together with a bidirectional orientation loss:
\begin{equation}
	\mathcal{L}_\theta = \frac{1}{HW} \sum_{y=1}^{H} \sum_{x=1}^{W} \Delta_\theta(\hat{\theta}(x,y),\theta(x,y))C_\theta(x,y), \quad \Delta_\theta(\theta_1, \theta_2) = \min(|\theta_1 - \theta_2|, \pi - |\theta_1 - \theta_2|)
	\label{eq:l_orien}
\end{equation}
where $\hat{\theta}$ and $\theta$ represents the rendered and precomputed orientation maps, and $C_\theta$ the corresponding confidence map. This loss accelerates convergence toward the correct hair orientation.

To further supervise $\alpha$, we introduce a mask loss based on the binary cross-entropy between the precomputed mask $M(x,y)$ and the rendered mask $\hat{M}(x,y)$:
\begin{equation}
	\mathcal{L}_{m} = - \frac{1}{HW} \sum_{y=1}^{H} \sum_{x=1}^{W} \left[ M(x, y) \log \left( \hat{M}(x,y) \right) + \left( 1 - M(x, y) \right) \log \left( 1 - \hat{M}(x,y) \right) \right]
	\label{eq:l_mask}
\end{equation}

The overall objective for the first-stage is:
\begin{equation}
	\mathcal{L}_{first} = (1-\lambda_{DSSIM})\mathcal{L}_1 + \lambda_{DSSIM}\mathcal{L}_{DSSIM} + \lambda_{\theta}\mathcal{L}_{\theta} + \lambda_{m}\mathcal{L}_{m}
	\label{eq:loss_first_stage}
\end{equation}

We use the default 3DGS densification mechanism to add Gaussians where geometry is missing, resulting in a dense and detailed scene representation for the next stages.

\subsection{Stage II: Strand Generation}
\label{sec:stage_2}

\noindent\textbf{Strand Representation.} To continue leveraging the 3DGS framework for strand optimization, we represent each hair strand as a chain of linked joints, similar to ~\cite{GroomCap, GS-Hair}. In contrast to the first stage, rasterization is now performed on the segments connecting these joints, rather than on the joints themselves. Each segment is modeled as a cylinder aligned with the x-axis, with its scale defined as $(||p_{j+1} - p_j||_2, \tau_j, \tau_j)$, where $\tau_j$ is a learnable thickness parameter. The center $\mu$ of each segment is set to the midpoint between its endpoints, and the rotation matrix $R$ is computed to align the x-axis vector $\vec{x} = (1,0,0)$ with the segment direction $\vec{p}j = \frac{p{j+1} - p_j}{|p{j+1} - p_j|_2}$ using the Rodrigues rotation formula:

\begin{equation}
	R = I + K + \frac{K^2}{1+v\cdot d}, \quad K = \vec{x} \times \vec{p}_j
	\label{eq:rodrigues}
\end{equation}
Other segment properties, such as opacity $\alpha$, mask $m$, and color $sh$, are optimized independently.

\noindent\textbf{Merging Scheme.} Initially, each Gaussian from the first stage is converted into a short strand with two joints, based on its position and direction. To form longer and more realistic (curvy) strands, we iteratively merge these short strands. This merging is formulated as an assignment problem, where strand endpoints (tips and roots) are nodes in a bipartite graph, and edge costs are determined by spatial distance and angular difference. While the Hungarian Matching Algorithm~\cite{Kuhn1955TheHM} can find the optimal solution, its computational and memory requirements are prohibitive for our large-scale problem (up to half a million nodes). Therefore, we use a more efficient greedy approach: candidates are filtered by predefined distance and angle thresholds, sorted by cost, and only the lowest-cost matches are kept. We implement this efficiently using a K-D tree-based nearest neighbor search.

For each selected pair, a new joint is created at the midpoint, and the new strand is formed by connecting the original endpoints to this joint. This process is illustrated in Strand Generation of Figure~\ref{fig:pipeline}.

To ensure high confidence in the initial merging process, we use very restrictive thresholds: $d_{m}=2$\,mm and $\theta_{m}=20^\circ$.

\subsection{Stage III: Growing and Refinement}
\label{sec:stage_3}

\noindent\textbf{Geometry Refinement.} After merging, we observe a gradual loss of geometric accuracy, mainly due to the naive placement of new joints at the midpoints of merged endpoints. To correct this, we refine joint positions by reintroducing image-based supervision with the losses from Stage 1.

Unconstrained optimization can also cause unnaturally sharp angles between connected segments. To prevent this, we add an angle smoothness loss that penalizes large direction changes. Let $\mathcal{C}$ be the set of all connected segment pairs $(a,b)$, $\vec{p}_a$, $\vec{p}_b$ their normalized direction vectors, and $\theta_{s}$ an angle threshold. The loss defined as:
\begin{equation}
	\label{eq:loss_smoothness}
	\mathcal{L}_{\text{smooth}} = \frac{1}{|\mathcal{C}|} \sum_{(a,b) \in \mathcal{C}} \theta_{a,b}^2, \quad
	\theta_{a,b} =
	\begin{cases}
		\cos^{-1}(\vec{p}_a \cdot \vec{p}_b) & \text{if } \vec{p}_a \cdot \vec{p}_b \leq \cos(\theta_{s}) \\
		0                                    & \text{otherwise}
	\end{cases}
\end{equation}
The overall loss for optimizing the joint positions of our Gaussian strands $\mathcal{GS}$ is:
\begin{equation}
	\label{eq:loss_third_stage}
	\mathcal{L}_{third}= \mathcal{L}_{first} +  \lambda_{smooth}\mathcal{L}_{smooth}
\end{equation}

\noindent\textbf{Topology Refinement.} Empirically, we find that sharp angles most often arise when modeling curly strands. While the smoothness loss helps prevent such artifacts, it alone is not sufficient to capture the complexity of curly hair. To address this, we split segments that exceed a length threshold, inserting new joints at their midpoints. This increases the degrees of freedom and allows the model to better represent curves. Subsequent image-based supervision further refines these new joint positions, resulting in more accurate strand shapes (see the Growing and Refinement stage in Figure~\ref{fig:pipeline}).

Despite these improvements, reconstructed strands may still be shorter than in reality or appear as multiple disconnected segments. This occurs because the merging algorithm cannot always identify endpoints belonging to the same underlying strand if they are not sufficiently close. To alleviate this problem, we gradually relax the merging criteria during optimization by increasing the distance and angle thresholds from $d_{m}=2mm$ and $\theta_{m}=20^\circ$ at the start, to $d_{m}=4mm$ and $\theta_{m}=40^\circ$ by the end. This strategy enables more merges and nearly doubles the average strand length.

Finally, we filter out non-hair geometries using the learned mask values. The final result is a clean set of hair strands represented as polylines, constructed from the optimized joint positions and connectivity.

\section{Experiments}
\label{sec:experiments}
\subsection{Experimental Setup}

\textbf{Datasets.} We evaluate our method and baselines on both synthetic and real-world datasets. For quantitative analysis, we use the USC-HairSalon~\cite{usc-hairsalon} dataset, which is, to our knowledge, the largest publicly available hair dataset, containing 343 hairstyles with 10,000 strands and 100 joints per strand. We also use the Cem Yuksel hair models~\cite{Yuksel2009} as a test set; these models are smaller in scale and not used for training by learning-based baselines. For each sample, we generate 16 input images by rotating a virtual camera around the subject in OpenGL, using simple ambient and diffuse lighting. Hair strands are rendered as line primitives together with the head mesh and assigned brown color with slight variations for realism.

For qualitative evaluation, we also test on the real-world NeRSemble dataset~\cite{kirschstein2023nersemble}, which features real head captures with diverse hairstyles.

\noindent\textbf{Baselines.} We compare our method against state-of-the-art 3D hair reconstruction approaches based on multiview images, including Neural Haircut~\cite{Neural-Haircut} as a representative data-driven method, and classical SfM-based methods LP-MVS~\cite{Strand-Accurate} and Strand Integration~\cite{Strand-Integration}. For LP-MVS, we use the implementation provided by the authors of Strand Integration, as no official code is available. Both methods return a directed point cloud without topological information, so we perform additional postprocessing to convert them into strands using the forward Euler method described in~\cite{Strand-Accurate}. However, this lacks the mean-shift and growing steps which are essential for producing clean, long hair strands. We therefore keep both versions and refer to the postprocessed results as LP-MVS$^\dag$ and Strand Integration$^\dag$.

\noindent\textbf{Implementation Details.} For the initial geometry reconstruction, we optimize the Gaussians for 30,000 iterations using the Adam optimizer~\cite{Adam}. The merging algorithm is run until no candidates remain, followed by an additional 30,000 iterations of refinement. All other hyperparameters are adopted directly from~\cite{GS}. Experiments are conducted on a NVIDIA RTX 4090 GPU and AMD Ryzen 9 7950X CPU, with the entire pipeline typically completing in about 1 hour. Our fully optimization-based approach is significantly faster than learning-based methods, which are generally more computationally intensive. For example, Neural Strands~\cite{Neural-Strands} takes 48 hours on a NVIDIA V100, Neural Haircut~\cite{Neural-Haircut} requires approximately 120 hours on our setup, and GroomCap~\cite{GroomCap} takes a similar amount of time on even higher-end hardware.

\subsection{Evaluation Measures}
Due to the difficulty of finding true correspondence between points. We follow common practices from\cite{Dr-Hair, Strand-Accurate, Neural-Haircut} to define precision, recall and f1-score based on matched nearest neighbors under a distance and angle threshold.

\noindent\textbf{Strand Consistency.} A key limitation of these metrics is that they only assess geometric accuracy at the point cloud level, ignoring the essential topological structure of hair strands. As long as the location of the points is correct, the metrics do not penalize even if the edges are completely wrong. To address this, we introduce a novel metric, strand consistency (SC), which evaluates the connectivity of points within hair strands. Specifically, the metric measures, for each Ground truth (GT) strand, the highest fraction of its points that are matched to points in a single predicted strand, and then averages this value over all strands.

Let $\mathcal{S}_G$ and $\mathcal{S}_P$ denote the sets of ground truth and predicted strands, respectively. And let $g$ and $p$ be points on $s_G$ and $s_P$. $\mathbf{I}(\cdot)$ is the indicator function and $d_{\tau}$ and $\theta_{\tau}$ are the distance and angle thresholds. The computation is as follows:
\begin{equation}
	\label{eq:strand_consistency}
	C = \frac{1}{|\mathcal{S}_G|} \sum_{s_G \in \mathcal{S}_G} \max_{s_P \in \mathcal{S}_P} \left( \frac{1}{|s_G|} \sum_{g \in s_G} \mathbf{I}\left( \exists\, p \in s_P : \|g - p\|_2 \leq d_{\tau} \;\wedge\; \theta_{g, p} \leq \theta_{\tau} \right) \right)
\end{equation}

Intuitively, this metric looks at each GT strand and finds the predicted strand with the most matching directed points.

\subsection{Comparison}
\begin{figure}[t]
	\centering
	\begin{minipage}[t]{0.77\textwidth}
		\centering
		\includegraphics[width=\textwidth]{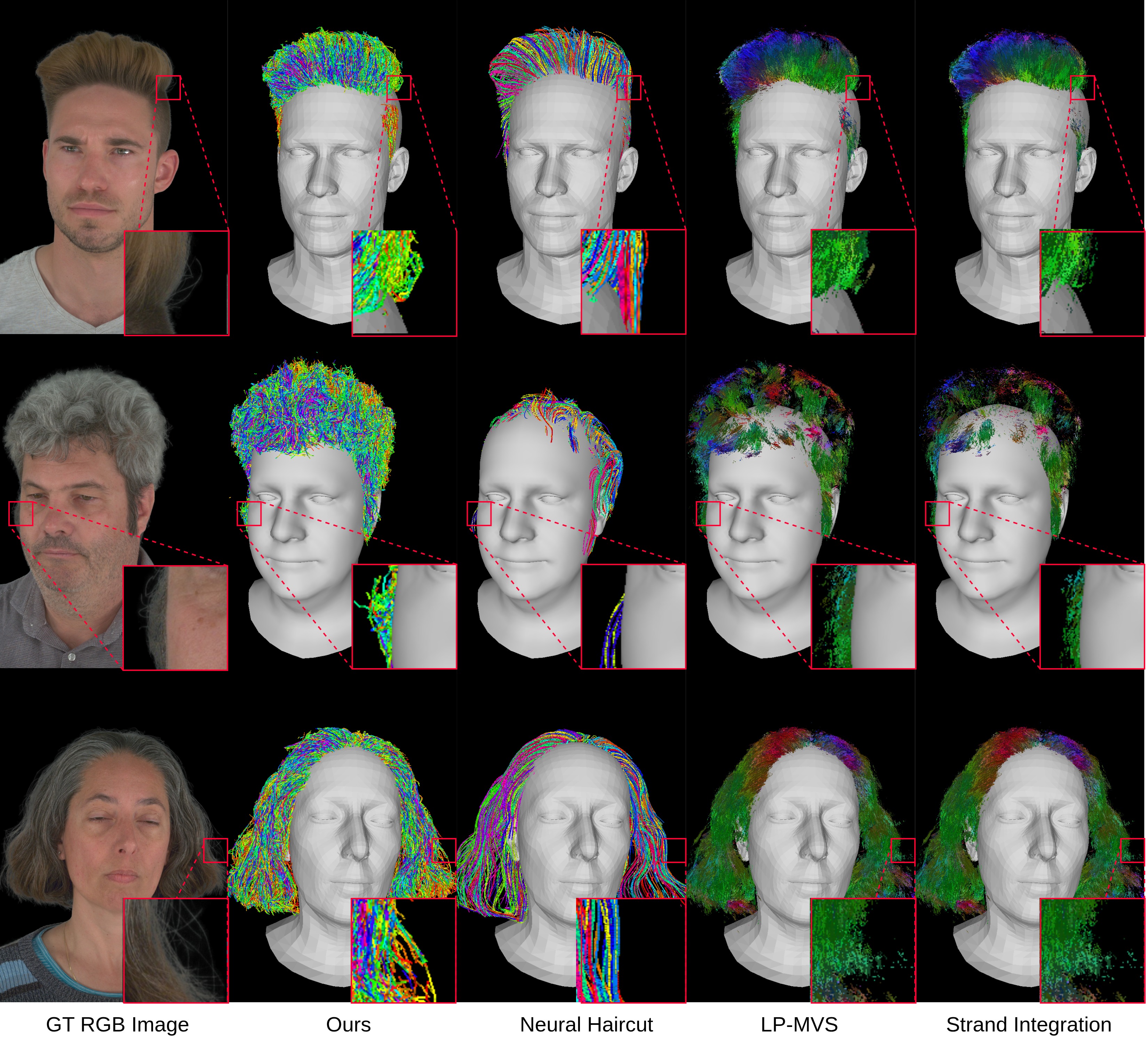}
		\caption{\textbf{Comparison on real-world subjects.} Our method excels on challenging hairstyles, especially curly hair. Colors either differentiate strands (Ours and Neural Haircut) or direction (others).}
		\label{fig:nersemble}
	\end{minipage}%
	\hfill
	\begin{minipage}[t]{0.2235\textwidth}
		\centering
		\includegraphics[width=\textwidth]{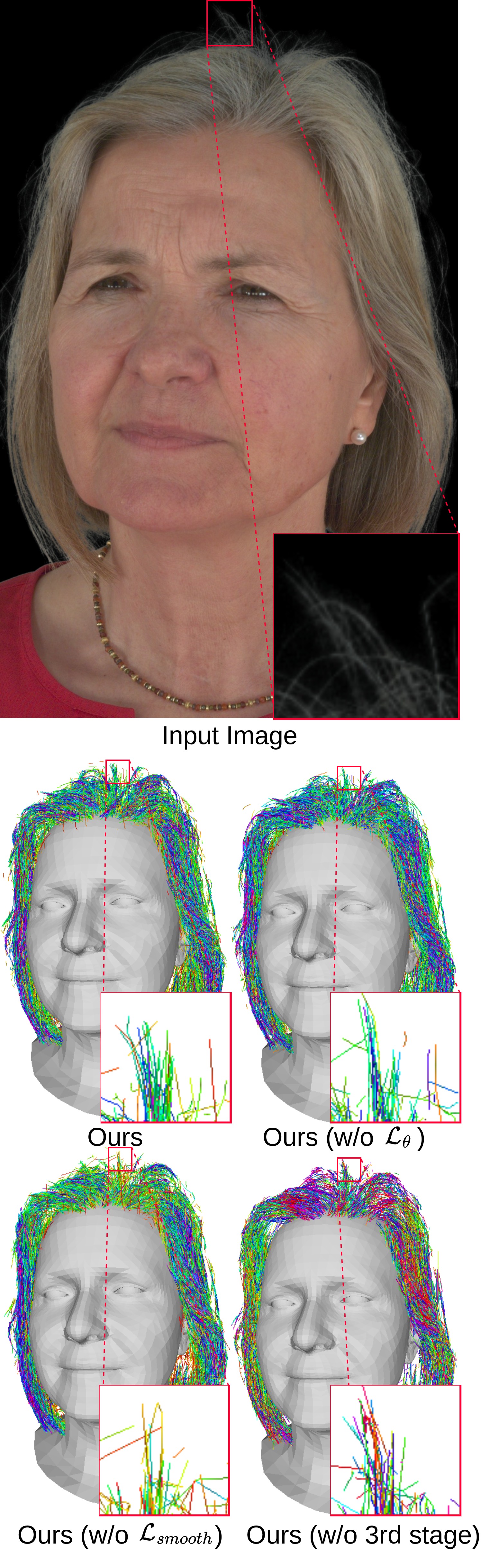}
		\caption{\textbf{Ablation study.} Full method gives best results.}
		\label{fig:ablation_study}
	\end{minipage}
\end{figure}

\noindent\textbf{Quantitative Results.} On the USC-HairSalon dataset, our method achieves higher precision but lower recall than Strand Integration, indicating that while MVS-based methods are robust to input variations, they often produce noisy results with incorrect directions due to reliance on hand-crafted filters (see Table~\ref{tab:usc_aggregated}). Neural Haircut shows consistently lower scores across all metrics, reflecting limited generalization to the synthetic evaluation data. In contrast, our fully optimization-based approach, which makes fewer assumptions from the input, demonstrates greater robustness and generalization. On the challenging curly hair sample from the Cem-Yuksel hair model, our method outperforms all others across every metric (Table~\ref{tab:wCurly}), and achieves the highest strand consistency, indicating more reliable reconstruction of correct strand connectivity.

\begin{table}
	\centering
	\tiny
	\setlength{\tabcolsep}{2.0pt} %
	\begin{minipage}{0.48\linewidth}
		\centering
		\begin{tabular}{l|cc|cc|cc|cc}
			\hline
			\multicolumn{1}{c|}{}              & \multicolumn{8}{c}{\textbf{Thresholds: mm / degree}}                                                                                                                                                                                      \\
			\cline{2-9}
			\textbf{Method}                    & \multicolumn{2}{c|}{\textbf{Precision}}              & \multicolumn{2}{c|}{\textbf{Recall}} & \multicolumn{2}{c|}{\textbf{F-score}} & \multicolumn{2}{c}{\textbf{SC}}                                                                     \\
			\cline{2-9}
			                                   & \textbf{2/20}                                        & \textbf{4/40}                        & \textbf{2/20}                         & \textbf{4/40}                   & \textbf{2/20}  & \textbf{4/40}  & \textbf{2/20}  & \textbf{4/40}  \\
			\hline
			\textbf{Ours}                      & 0.445                                                & \textbf{0.769}                       & 0.182                                 & \textbf{0.545}                  & 0.256          & \textbf{0.636} & \textbf{0.050} & \textbf{0.189} \\
			\textbf{Neural Haircut}            & 0.187                                                & 0.603                                & 0.040                                 & 0.192                           & 0.064          & 0.291          & 0.023          & 0.094          \\
			\textbf{LP-MVS}                    & 0.388                                                & 0.643                                & 0.209                                 & 0.505                           & 0.271          & 0.565          & --             & --             \\
			\textbf{LP-MVS$^\dag$}             & 0.229                                                & 0.474                                & 0.098                                 & 0.345                           & 0.136          & 0.398          & 0.019          & 0.046          \\
			\textbf{Strand Integration}        & \textbf{0.497}                                       & 0.712                                & \textbf{0.243}                        & 0.533                           & \textbf{0.326} & 0.608          & --             & --             \\
			\textbf{Strand Integration$^\dag$} & 0.258                                                & 0.486                                & 0.117                                 & 0.384                           & 0.161          & 0.427          & 0.022          & 0.053          \\
			\hline
		\end{tabular}
		\caption{Quantitative comparison on aggregated samples from the USC-HairSalon dataset. Higher values indicate better results and are marked in bold.}
		\label{tab:usc_aggregated}
	\end{minipage}
	\hfill
	\begin{minipage}{0.48\linewidth}
		\centering
		\begin{tabular}{l|cc|cc|cc|cc}
			\hline
			\multicolumn{1}{c|}{}              & \multicolumn{8}{c}{\textbf{Thresholds: mm / degree}}                                                                                                                                                                                      \\
			\cline{2-9}
			\textbf{Method}                    & \multicolumn{2}{c|}{\textbf{Precision}}              & \multicolumn{2}{c|}{\textbf{Recall}} & \multicolumn{2}{c|}{\textbf{F-score}} & \multicolumn{2}{c}{\textbf{SC}}                                                                     \\
			\cline{2-9}
			                                   & \textbf{2/20}                                        & \textbf{4/40}                        & \textbf{2/20}                         & \textbf{4/40}                   & \textbf{2/20}  & \textbf{4/40}  & \textbf{2/20}  & \textbf{4/40}  \\
			\hline
			\textbf{Ours}                      & \textbf{0.411}                                       & \textbf{0.736}                       & \textbf{0.139}                        & \textbf{0.468}                  & \textbf{0.207} & \textbf{0.572} & \textbf{0.041} & \textbf{0.166} \\
			\textbf{Neural Haircut}            & 0.190                                                & 0.639                                & 0.008                                 & 0.066                           & 0.015          & 0.120          & 0.006          & 0.042          \\
			\textbf{LP-MVS}                    & 0.302                                                & 0.532                                & 0.052                                 & 0.289                           & 0.089          & 0.375          & --             & --             \\
			\textbf{LP-MVS$^\dag$}             & 0.212                                                & 0.472                                & 0.033                                 & 0.226                           & 0.056          & 0.306          & 0.015          & 0.064          \\
			\textbf{Strand Integration}        & 0.154                                                & 0.346                                & 0.021                                 & 0.136                           & 0.037          & 0.195          & --             & --             \\
			\textbf{Strand Integration$^\dag$} & 0.106                                                & 0.300                                & 0.012                                 & 0.107                           & 0.022          & 0.157          & 0.007          & 0.038          \\
			\hline
		\end{tabular}
		\caption{Quantitative comparison on a challenging curly hair sample from the Cem-Yuksel hair model. Higher values indicate better results.}
		\label{tab:wCurly}
	\end{minipage}
\end{table}

\noindent\textbf{Qualitative Results.} Further analysis on the NeRSemble dataset~\cite{kirschstein2023nersemble} (Figure~\ref{fig:nersemble}) shows results for three hairstyles: straight, curly, and long hair. Neural Haircut improves on real-world data, generally matching the overall shape for straight hair. However, only our method accurately captures fine details, such as thin and floating strands in the female subject. For curly hair, which is especially challenging due to rapid directional changes, LP-MVS and its variants capture only the sparse outer geometry, while Neural Haircut fails to represent the overall shape. In contrast, our optimization-based method, free from prior assumptions, robustly reconstructs all tested hairstyles.

\subsection{Ablation Study}

\begin{table}[h]
	\centering
	\setlength{\tabcolsep}{3.0pt} %
	\begin{tabular}{l|cc|cc|cc|cc}
		\hline
		\multicolumn{1}{c|}{}                            & \multicolumn{8}{c}{\textbf{Thresholds: mm / degree}}                                                                                                                                                                                      \\
		\cline{2-9}
		\textbf{Merging thresholds}                      & \multicolumn{2}{c|}{\textbf{Precision}}              & \multicolumn{2}{c|}{\textbf{Recall}} & \multicolumn{2}{c|}{\textbf{F-score}} & \multicolumn{2}{c}{\textbf{SC}}                                                                     \\
		\cline{2-9}
		                                                 & \textbf{2/20}                                        & \textbf{4/40}                        & \textbf{2/20}                         & \textbf{4/40}                   & \textbf{2/20}  & \textbf{4/40} & \textbf{2/20}  & \textbf{4/40}   \\
		\hline
		\textbf{$1mm/10^\circ \rightarrow 2mm/20^\circ$} & 0.4264                                               & 0.7606                               & 0.1281                                & 0.4819                          & 0.197          & 0.59          & 0.048          & 0.1822          \\
		\textbf{$2mm/20^\circ \rightarrow 4mm/40^\circ$} & \textbf{0.4639}                                      & 0.8037                               & 0.15                                  & \textbf{0.54}                   & 0.2277         & 0.6438        & \textbf{0.055} & 0.2196          \\
		\textbf{$4mm/40^\circ \rightarrow 6mm/60^\circ$} & 0.4618                                               & \textbf{0.8213}                      & \textbf{0.155}                        & 0.5383                          & \textbf{0.233} & \textbf{0.65} & 0.055          & \textbf{0.2262} \\
		\textbf{$6mm/60^\circ \rightarrow 8mm/80^\circ$} & 0.4355                                               & 0.775                                & 0.1376                                & 0.5182                          & 0.271          & 0.565         & 0.05           & 0.1866          \\
		\hline
	\end{tabular}
	\caption{Impact of different initial and final merging thresholds on reconstruction performance. Results show that performance remains stable within a reasonable range, while overly extreme thresholds degrade accuracy. Bold numbers indicate the best results.}
	\label{tab:threshold_ablation}
\end{table}

We evaluate the impact of key design choices in our method (Figure~\ref{fig:ablation_study}). Removing the angle smoothness loss leads to unrealistic, spiky strands due to the lack of regularization on segment directions. Skipping the third-stage optimization, which refines joint positions and adaptively inserts points, produces coarse and inaccurate strands, as the initial merging only yields rough approximations. Restrictive merging criteria in the second stage prevent short strands from being combined into longer ones, while enforcing a limited number of joints constrains the representation to short, straight segments. Without orientation supervision, hair segments often deviate in incorrect directions. In contrast, the full method integrates all components and achieves the most faithful strand reconstructions.

We further analyze the impact of different merging thresholds (Table~\ref{tab:threshold_ablation}). As shown in the first two rows, varying the thresholds within a reasonable range leads to only minor performance differences. A noticeable drop in accuracy occurs only when the thresholds are pushed to extremes—either overly restrictive or overly permissive, as in the first and fourth rows—indicating that our method remains robust to threshold selection as long as it stays within practical bounds.

\section{Conclusion}
\label{sec:conclusion}

In this work, we introduce a fully optimization-based method for hair reconstruction from multi-view images, capable of handling a wide range of hairstyles without relying on priors from synthetic datasets. Our multi-stage pipeline first estimates an accurate Gaussian representation using 3DGS, then merges individual segments to form complete hair strands.

To address the limitations of existing evaluation metrics, which overlook the structural connectivity of hair strands, we propose a novel metric as a proxy for topological accuracy.

Extensive experiments on both synthetic and real-world datasets show that our method outperforms existing approaches, producing highly detailed hair models—including fine, floating strands—while being considerably faster than data-driven methods, with reconstructions completed in under one hour.

Beyond hair, our framework could naturally extend to other line-like structures such as cables or wires, requiring minimal adaptation of the segmentation model.

Despite these strengths, some limitations remain. Our merging criteria can prevent effective merging of Gaussians from the same strand, resulting in shorter reconstructed strands. As discussed before using a variation of Hungarian algorithm for optimal matching could address this. Additionally, reconstructed strands are not necessarily attached to the scalp, limiting their use in rendering engines; future work could address this by pinning strand roots to the surface and growing strands as in~\cite{MonoHair, GroomCap}.

\section{Acknowledgements}
\label{sec:acknowledgements}

This work was supported by the ERC Consolidator Grant Gen3D (101171131) and the German Research Foundation (DFG) Research Unit “Learning and Simulation in Visual Computing”.

\bibliography{egbib}
\end{document}